\documentclass[runningheads]{llncs}
\usepackage{graphicx}

\usepackage{tikz}
\usepackage{comment}
\usepackage{amsmath,amssymb} 
\usepackage{color}

\usepackage[accsupp]{axessibility}  

\usepackage{graphicx}
\usepackage{amsmath}
\usepackage{amssymb}
\usepackage{booktabs}
\usepackage{bm}
\usepackage{multirow}
\usepackage{adjustbox}
\usepackage{booktabs}
\usepackage{lipsum}
\usepackage{color}
\usepackage{verbatim}
\usepackage{caption}
\usepackage{makecell}
\usepackage{arydshln}

\usepackage{colortbl,hhline}
\usepackage{tabulary}
\usepackage{etoolbox}

\usepackage{cite}

\newcommand\blfootnote[1]{%
  \begingroup
  \renewcommand\thefootnote{}\footnote{#1}%
  \addtocounter{footnote}{-1}%
  \endgroup}
\usepackage[pagebackref,breaklinks,colorlinks]{hyperref}
\usepackage[capitalize]{cleveref}
\crefname{section}{Sec.}{Secs.}
\Crefname{section}{Section}{Sections}
\Crefname{table}{Table}{Tables}
\crefname{table}{Tab.}{Tabs.}

\begin{document}

\pagestyle{headings}
\mainmatter
\def\ECCVSubNumber{100}  

\title{DevNet: Self-supervised Monocular Depth Learning via Density Volume Construction} 

\titlerunning{DevNet: Monocular Depth Learning via Density Volume Construction}
\author{Kaichen Zhou\inst{1} 
\and
Lanqing Hong\inst{2\dagger}
\and
Changhao Chen\inst{1}
\and
Hang Xu\inst{2} 
\and
Chaoqiang Ye\inst{2} 
\and
Qingyong Hu\inst{1} 
\and
Zhenguo Li\inst{2}}
\authorrunning{K. Zhou et al.}
%
\institute{University of Oxford \and Huawei Noah’s Ark Lab \\
\email{kc.zhou2020@hotmail.com},
\email{\{honglanqing,yechaoqiang,xu.hang,li.zhenguo\}@huawei.com},
\email{\{changhao.chen66,huqingyong15\}@outlook.com}}
\maketitle

\begin{abstract}
Self-supervised depth learning from monocular images normally relies on the 2D pixel-wise photometric relation between temporally adjacent image frames. 
However, they neither fully exploit the 3D point-wise geometric correspondences, 
nor effectively tackle the ambiguities in the photometric warping caused by occlusions or illumination inconsistency.
To address these problems, this work proposes Density Volume Construction Network (DevNet), a novel self-supervised monocular depth learning framework, that can consider 3D spatial information, and exploit stronger geometric constraints among adjacent camera frustums. 
Instead of directly regressing the pixel value from a single image, our DevNet divides the camera frustum into multiple parallel planes and predicts the pointwise occlusion probability density on each plane. The final depth map is generated by integrating the density along corresponding rays. 
During the training process, novel regularization strategies and loss functions are introduced to mitigate photometric ambiguities and overfitting.
Without obviously enlarging model parameters size or running time, DevNet outperforms several representative baselines on both the KITTI-2015 outdoor dataset and NYU-V2 indoor dataset. 
In particular, the root-mean-square-deviation is reduced by around 4\% with DevNet on both KITTI-2015 and NYU-V2 for the unsupervised monocular depth estimation task.
\blfootnote{$\dagger$ Corresponding author.}
\end{abstract}

\section{Introduction}\label{sec:intro}
\begin{figure}[t]
\setlength{\abovecaptionskip}{0.1cm}
\setlength{\belowcaptionskip}{-0.5cm}
    \centering
    \includegraphics[width=11.5cm]{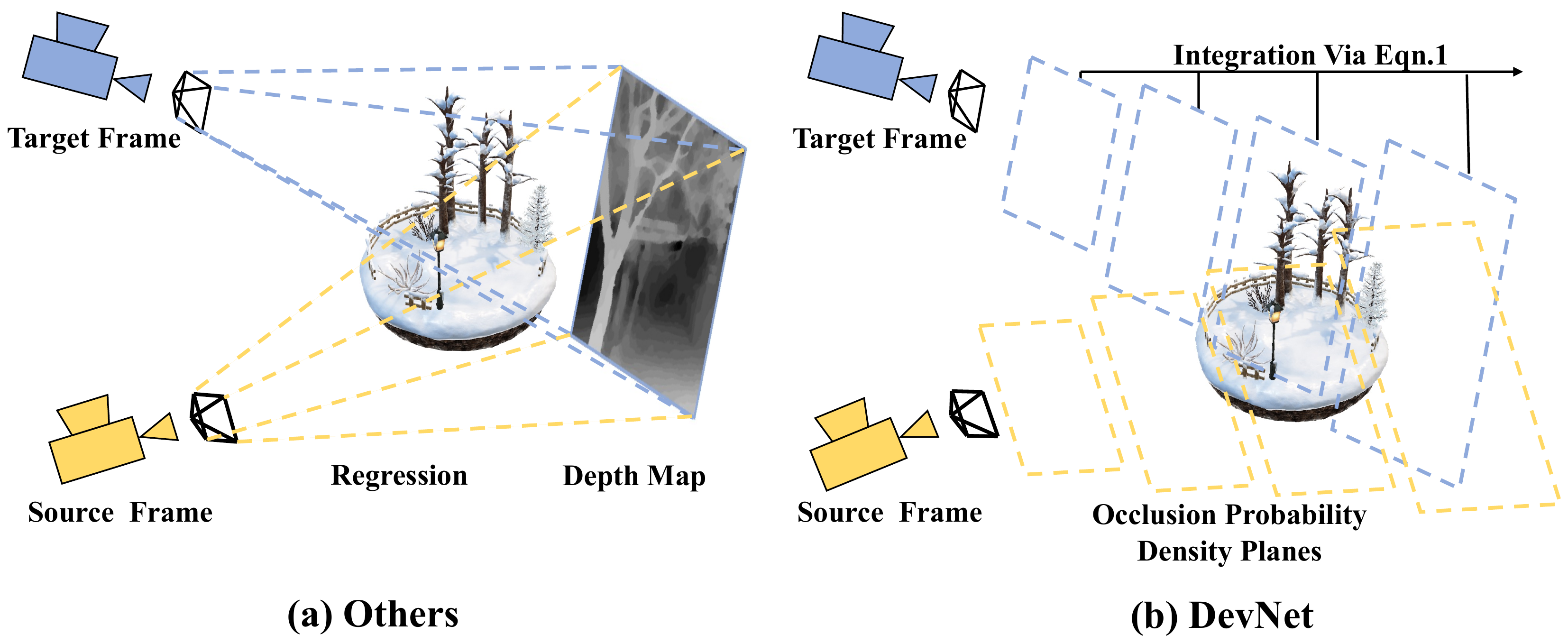}
    \caption{\textbf{A comparison between our proposed DevNet and other depth learning approaches \cite{pillai2019superdepth, gordon2019depth, guizilini20203d, shu2020feature, zhou2021r}.} \textbf{(a)} Most existing depth learning methods predict depth maps directly from input images \cite{godard2017unsupervised, godard2019digging}, and the correspondences among adjacent frames are built only with 2D pixel-wise depth maps. \textbf{(b)} Our DevNet renders depth maps by predicting the density information of parallel planes, and the correspondences among adjacent frames are built with 3D frustums. The pyramid denotes the frustum.}
    \label{fig:fig1}
\end{figure}

Vision-based depth estimation (VDE) attracts attentions due to its significance in understanding the geometry of a 3D scene.  It is the basis of higher-level 3D tasks, e.g., scene reconstruction \cite{liu2020depth} and object detection \cite{sun2020disp}, and supports a number of cutting-edge applications, from autonomous driving \cite{huo2020learning, zhang2020data} to augmented reality \cite{kalia2019real}.  

Recently learning-based VDE becomes a focus, that can be generally divided into two categories -- supervised  \cite{bhat2021adabins, eigen2015predicting} and self-supervised approaches \cite{zhan2018unsupervised, zhou2017unsupervised, zhu2020edge}. Supervised VDE usually requires high-precision ground-truth depth as training labels, whose process would be costly and time-consuming. Self-supervised methods learn depth maps from monocular images \cite{zhou2017unsupervised} or stereo image pairs \cite{zhu2020edge}. 
Self-supervised monocular VDE \cite{godard2019digging, shu2020feature, zhou2021r}, exploits the photometric loss based on the warping relation between temporally adjacent image frames \cite{zhou2017unsupervised}. 
Along with per-pixel depth predictions, it also produces ego-motion estimation.
Though self-supervised monocular VDE has seen great progresses, there still exists a large performance gap between self-supervised and supervised methods \cite{peng2021excavating}. 
This gap comes from the fact that it is unreliable to optimize the framework only based on the photometric loss without understanding the geometry of the whole scene. The assumption that lower photometric loss denotes higher depth accuracy does not hold in many cases, e.g., the occlusions, illumination inconsistency, non-Lambertian surfaces, and textureless regions.

To tackle these problems, we propose DevNet, a novel density volume construction based monocular depth learning framework. \textit{By exploiting volume rendering \cite{kajiya1986rendering}, 3D spatial information along the ray can be integrated into pixel-wise depth estimation, which contributes to building stronger correspondences among adjacent camera frustums.}
Instead of directly producing a depth map from a single RGB image, DevNet discretizes the scene into multiple parallel planes, as shown in Fig.~\ref{fig:fig1}(b), and predicts the occlusion probability density information of each point on planes. 
Then, the pixelwise depth value is computed by integrating the probability density values of intersections between the ray and planes. 
During the training phase, our DevNet mitigates the limitations of photometric loss by adopting an occlusion regularization and a brightness regularization. Finally, a depth consistency loss based on occupancy probability density volume is introduced to maintain the geometric consistency between temporally adjacent frames. 
Extensive experiments conducted on both the KITTI-2015 \cite{geiger2012we} and NYU-V2 \cite{silberman2012indoor} datasets demonstrate the effectiveness of our proposed DevNet. 

Our main contributions are summarized as below:
\begin{itemize}
    \item We introduce DevNet, a novel self-supervised depth learning framework that renders depth maps via predicting the occlusion probability density of sampled points in the scene.
    
     \item We present a novel depth consistency loss using the predicted density volume, to ensure the geometric consistency between the depth maps of adjacent frames, which can further contribute to performance improvement.
    
    \item An occlusion regularization is introduced to detect and reduce the photometric ambiguity caused by camera motion and dynamic objects. 
\end{itemize}

\section{Related Works}

\subsection{Supervised Monocular Depth Learning}
In monocular depth learning, supervised approaches \cite{eigen2015predicting,zhou2019unsupervised,lee2019big,peng2021excavating} usually take a single image as input and use depth data measured with range sensors (e.g. RGB-D camera or LIDAR) as ground truth labels during training.
\cite{saxena2005learning} first proposed a learning-based method to learn depth from single monocular images in an end-to-end manner and the final algorithm could recover even accurate prediction for unstructured scenes. With the development of deep learning, \cite{eigen2015predicting} constructs a monocular depth estimator based on the convolutional architecture to infer the corresponding depth with a global estimation layer, followed by a local refinement layer. \cite{reading2021categorical} further improved neural network-based depth estimation by using a predicted categorical depth distribution for each pixel. 
Although the depth estimation network trained with ground-truth labels can achieve high accuracy in this task, obtaining high-resolution ground-truth labels from different domain/scenes is costly, time-consuming and not always applicable, which limits the potentials of these supervised methods.

\subsection{Self-supervised Depth Learning}
Self-supervised methods realize depth learning based on a series of unlabeled images. \cite{2016unsupervised} firstly made use of a pair of images to estimate corresponding by minimizing a photometric loss between the left input image and the warped right image as the target of optimization. \cite{zhou2017unsupervised} stated that only based on image reconstruction loss would lead to a poor result. They further improved the performance of depth estimation by proposing a new network architecture and a novel left-right consistency loss. 
However, due to the convenience of self-supervised monocular depth learning, it attracts more attention. \cite{zhou2017unsupervised} designed an end-to-end self-supervised monocular depth estimation network for depth estimation and motion estimation of monocular unlabeled video. By using the estimated pose, the loss is constructed based on warping nearby views to the target view. \cite{yin2018geonet} further improved the accuracy of the self-supervised depth estimation task by jointly optimizing depth estimation, optic flow, and camera ego-motion. Predictions from three modules are used to construct the target view and form the loss function for both static background and dynamic components. \cite{godard2019digging} proposes lots of useful strategies such as a robust reprojection loss, multi-scale sampling, and an automasking loss to further improve the accuracy of monocular depth estimation. To improve the performance at non-contiguous regions and motion boundaries.
\cite{bian2019unsupervised} found that image reconstruction loss ignores the influence of dynamic components and proposes to exploit the geometric consistency among different frames for realizing scale-consistency prediction results. 
\cite{johnston2020self} further introduces the self-attention mechanism and takes the discrete disparity prediction into the framework for monocular depth estimation, which could learn more general contextual information and generate a more robust prediction. 

\subsection{Neural Rendering}
Rendering represents the process of generating 2D images from a 2D/3D model. Classic rendering contains two common approaches: Rasterization and Raytracing. Recent neural rendering generally denotes the combination of the recent proposed generative model with traditional computer graphics rendering techniques, which generates images of 3D scenarios from different camera views based on learned implicit representation. Based on this conception, NeRF \cite{mildenhall2020nerf} is proposed to encode both the appearance and the geometry information of a 3D scenario with a Multilayer Perceptron (MLP), which realizes the view synthesis of real-life scenes.
After that, numerous studies further adapt NERF to various scenarios, including images in wild~\cite{MartinBrualla20arxiv_nerfw}, dynamic scenes with monocular or stereo videos~\cite{pumarola2021d,li2021neural,du2021neural}, post estimation~\cite{yen2020inerf}, and depth estimation~\cite{wei2021nerfingmvs}. Despite diverse applications of neural rendering, we are the first to integrate this technique in the monocular depth estimation process. Considering the characteristics of this task and neural rendering, we further propose a novel occlusion regularization and depth consistency loss function to improve its performance in depth learning. 

\begin{figure*}[t]
\setlength{\abovecaptionskip}{0.1cm}
\setlength{\belowcaptionskip}{-0.5cm}
    \centering
    \includegraphics[width=12.0cm]{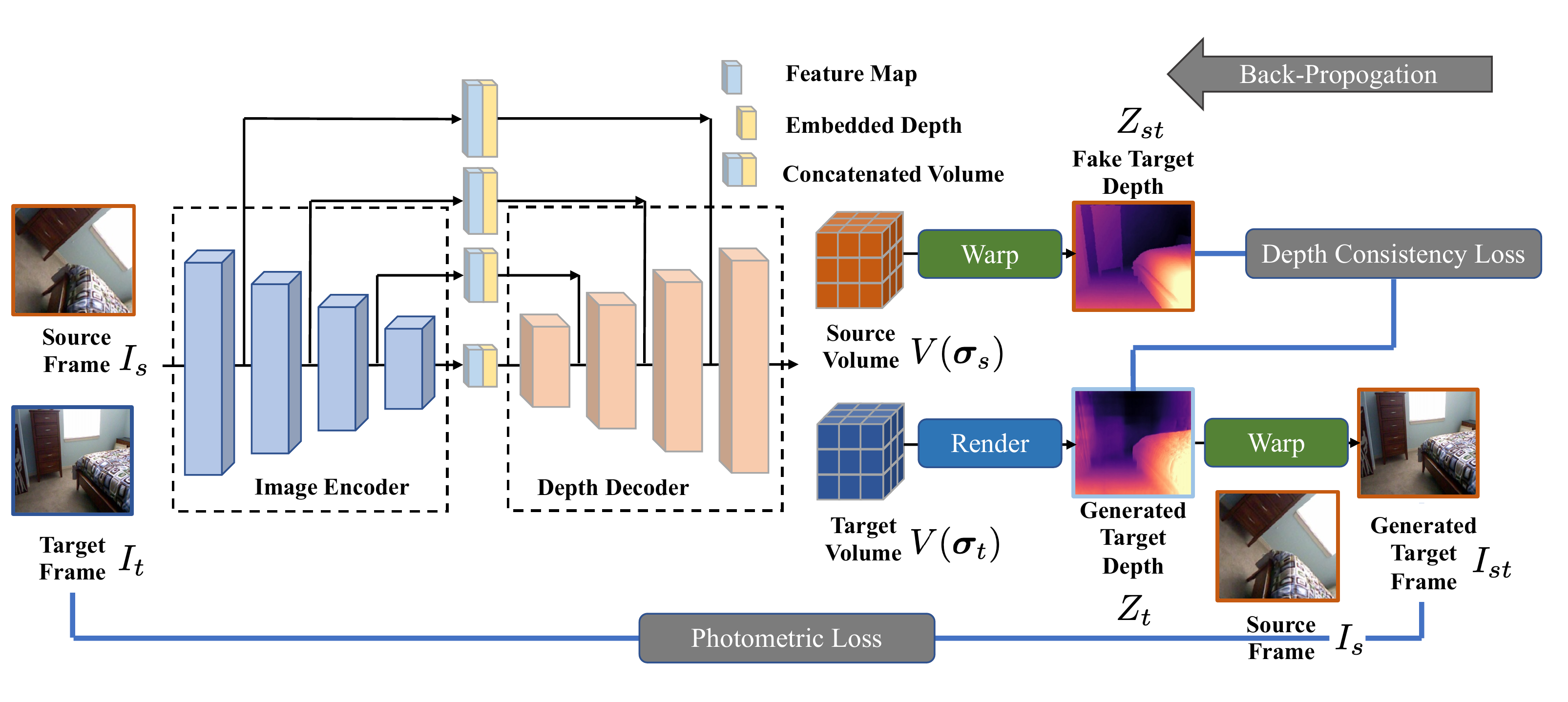}
    \caption{\textbf{The illustration for the Rendering-Based Depth Module}. This module is designed for estimating a density volume from an RGB frame and the density volume would be used to render the depth map. During the training phase, the photometric loss and depth consistency loss are used to realize self-supervised learning.}
\label{fig:fig2}
\end{figure*} 

\section{Density Volume Rendering Based Depth Learning}
\label{sec:method}
Most previous self-supervised monocular VDEs \cite{pillai2019superdepth, gordon2019depth, guizilini20203d, shu2020feature, zhou2021r} adopt the encoder-decoder structure to predict depth map directly from a single RGB image. 
However, it is not always reliable to optimize these models only by minimizing the photometric loss constructed by warping adjacent frames as in Fig.\ \ref{fig:fig1}(a) without considering the 3D spatial constraints.
DevNet aims to build stronger geometric constraints among adjacent frames by learning the density volume of corresponding camera frustums. 
The framework of DevNet consists of a rendering-based depth module and a pose module, as introduced in Sec.\ \ref{sec3:pipeline}. To reduce the influences of occlusions and illumination inconsistency, an occlusion regularization and a brightness regularization are introduced in Sec.\ \ref{sec3:mask}. We achieve the self-supervised depth estimation via using multiple loss functions in Sec.\ \ref{sec3:self-supervised}. Finally, Sec.\ \ref{sec3:discussion} discusses the differences between proposed DevNet and several similar works.

\subsection{Preliminary}
Volume rendering has been widely used in computing 2D RGB projections based on discrete 3D samples. Based on \cite{max1995optical}, to compute the color information along the ray $\bm{r} = \bm{o} + \bm{d} * s$ within the depth range $[s_n, s_f]$, we divide it into $K$ intervals with the length $\delta_{k}$ for each interval. The color information can be approximated as
\begin{equation}
    C = \sum^{K}_{k=1} T_k (1 - \exp(-\sigma_{k}\delta_{k}))c_k, \quad \text{where} \quad T_k = \exp(-\sum^{k-1}_{k'=1}\sigma_{k'}\delta_{k'}),
\label{eqn:render_color}
\end{equation}
where $\sigma_{k}$ denotes occlusion probability density at interval $\delta_k$; the occlusion probability within $\delta_{k}$ is written as $\sigma_{k}\delta_{k} = 1 - \exp(-\sigma_{k}\delta_{k})$,  when $\delta_{k}$ is small enough. Eqn.\ \eqref{eqn:render_color} can be regarded expectation of color information $c(s)$ in the occlusion distribution $\sigma_k$. Similar to color information, the depth is computed as
\begin{equation}
    Z = \sum^{K}_{k=1} T_k (1 - \exp(-\sigma_{k}\delta_{k}))z_k,
\label{eqn:render_depth}
\end{equation}
where $z_k$ is the depth value for point $s_k$ with respect to image plane. With Eqn.\ \eqref{eqn:render_depth}, the depth computation can exploit the spatial correspondence among different frames. This advantage has already been proven in \cite{mildenhall2020nerf,wei2021nerfingmvs}.

\subsection{Framework}\label{sec3:pipeline}
\begin{figure}[t]
\setlength{\abovecaptionskip}{-0.1cm}
\setlength{\belowcaptionskip}{-0.3cm}
    \centering
    \includegraphics[width=10cm]{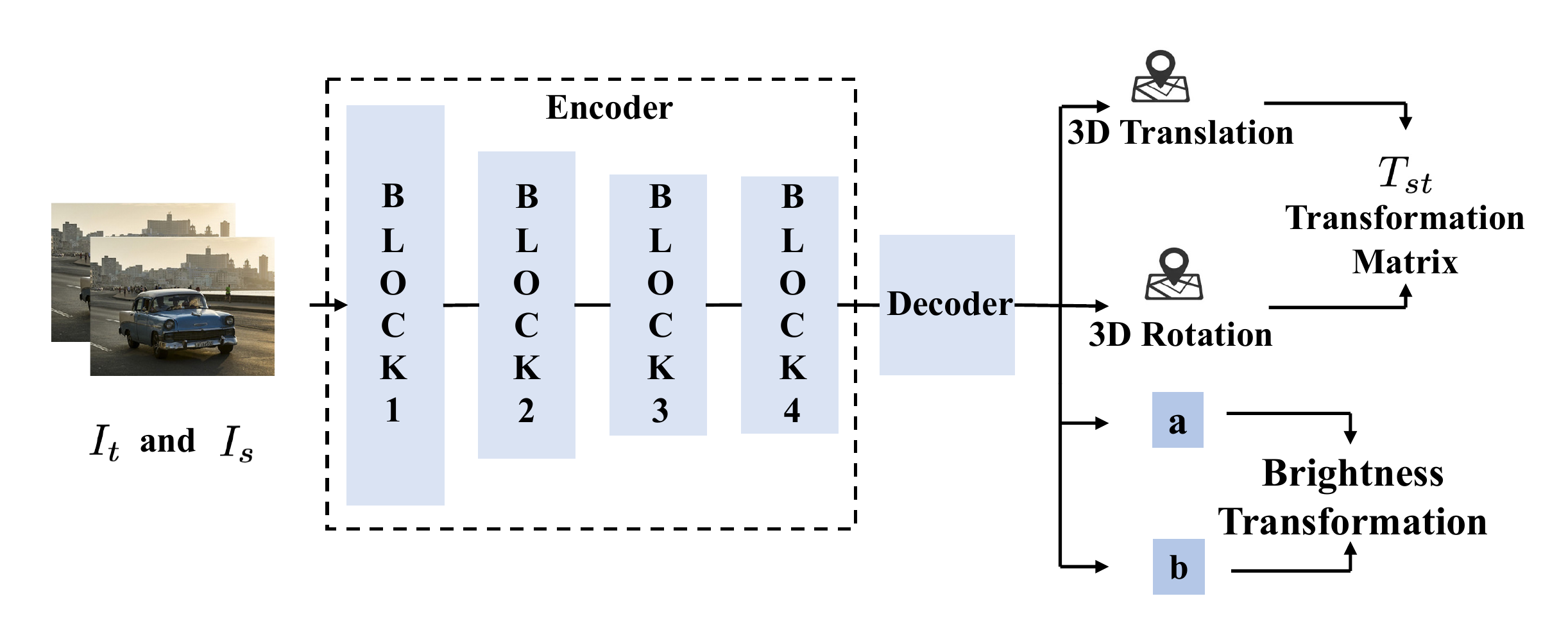}
    \caption{\textbf{The architecture of the Pose Module}. This module would take concatenated two frames as input and output the pose transformation and the brightness transformation parameters.}
\label{fig:fig3}
\end{figure}

DevNet framework contains a depth module for the occlusion probability density volume construction in Fig.\ \ref{fig:fig2}, and a pose module predicting the motion transformation between the target frame and source frame in Fig.\ \ref{fig:fig3}.
Specifically, DevNet discretizes the scene into multiple parallel planes as in Fig.\ \ref{fig:fig1}(b) and predicts the occlusion probability density of each point on planes. 
The pixelwise depth value is calculated through rendering over the ray passing through both the camera origin and the pixel location. 

\noindent\textbf{Density Volume Prediction:}
Given a RGB image $I_i \in \mathbb{R}^{C \times H \times W}$, instead of directly producing the depth $Z_i$ as in the previous research (Fig.\ \ref{fig:fig1}(a)), DevNet divides the camera frustum into $K$ multiple parallel planes $\{ \Pi_i^k \}_{k=1}^K$ located at the depth $\{ z_i^k \}_{k=1}^K$ (Fig.\ \ref{fig:fig1}(b)). Our depth module produces the occlusion probability  density of planes $V(\bm{\sigma}_i) = \{ \bm{\sigma}_i^k \}_{k=1}^K, \bm{\sigma}_i^k \in \mathbb{R}^{1 \times H \times W}$ based on  $I_i$. Finally, the $Z_i$ map is predicted through volume rendering Eqn.\ \eqref{eqn:render_depth}.

\noindent\textbf{Rendering-Based Depth Estimation Module:} The framework of depth module is shown in Fig.\ \ref{fig:fig2}, including an image encoder and a depth decoder. The image encoder is based on the ResNet structure \cite{he2016deep} which takes $I_i$ as input and generates hierarchical feature maps $\{ \bm{f}_i^j \}_{j=1}^N, \bm{f}_i^j \in \mathbb{R}^{C^j \times \frac{H}{2^{j}} \times \frac{W}{2^{j}}}$, where we adopt $N=4$ in our experiment. The the embedding strategy \cite{mildenhall2020nerf} is introduced to process depth information of each plane, written as
\begin{equation}
    e(z_i^k) = (\sin(2^0\pi z_i^k), \cos(2^0\pi z_i^k), ..., \sin(2^{(\frac{E}{2}-1)}\pi z_i^k), \cos(2^{(\frac{E}{2}-1)}\pi z_i^k)),
\end{equation}
where $E$ is dimension of embedded depth. After that, feature maps are concatenated with the corresponding embedded depth to obtain multi-scale feature maps $\{\text{cat}(e(z_i)^j, \bm{f}_i^j)\}_{j=1}^N$. 
These feature maps are skip-connected with the decoder to generate the probability density volume $V(\bm{\sigma}_i)$. During the training phase, DevNet outputs multi-scale density volumes $V(\bm{\sigma}_i) = \{ V^{j}(\bm{\sigma}_i) \}_{j=1}^N$ all of which are used in constructing the training loss. During the inference stage, DevNet only uses the biggest layer density volume.  

\noindent\textbf{Pose Estimation Module:}
This module also adopts encoder-decoder structure as in Fig.\ \ref{fig:fig3}. The pose encoder adopts the ResNet structure and takes concatenated target frame $I_{t}$ and source frame $I_{s}$ as input. The pose decoder takes the feature map from the last layer of the encoder as input. It outputs 3-dimensional translation and 3-dimensional rotation vectors, which are used to construct the transformation matrix $T_{ts}$, and the 2-dimensional brightness transformation parameters presented in Sec.\ \ref{sec3:mask}.

\subsection{Regularizations}\label{sec3:mask}
The self-supervised depth learning algorithms \cite{godard2017unsupervised, godard2019digging, yang2020d3vo, shu2020feature} generally rely on the warping relation between the pixel $P_{s}^{m}(x_{s}^{m},y_{s}^{m},1)$ in the source frame and its counterpart $P_{st}^{m}(x_{t}^{m},y_{t}^{m},1)$ in the target frame, which can be written as 
\begin{equation}
    P_{st}^m = Z_{t}(P_{st}^m)^{-1} K_{t} T_{st} Z_s(P_s^m) K^{-1}_{s} P_{s}^m,
\label{eqn:warp}
\end{equation}
where $K_{t}, K_{s}$ are the intrinsic matrices for the target frame and the source frame; $T_{st}$ is the transformation matrix from the source frame to the target frame; $Z_s(P_s^m)$ is the depth of pixel $P^m_s$ in source camera frustum and $Z_{t}(P_{ts}^m)$ is the depth of pixel $P^m_{ts}$ with the respect to the target camera frustum. However, this relation is destroyed in the following cases, such as moving objects, the camera motion, the illumination inconsistency, non-Lambertian surfaces, etc. To mitigate these influences of aforementioned occasions, we propose the following regularization strategies.

\begin{figure}[t]
\setlength{\abovecaptionskip}{0.1cm}
\setlength{\belowcaptionskip}{-0.65cm}
    \centering
    \includegraphics[width=9.0cm]{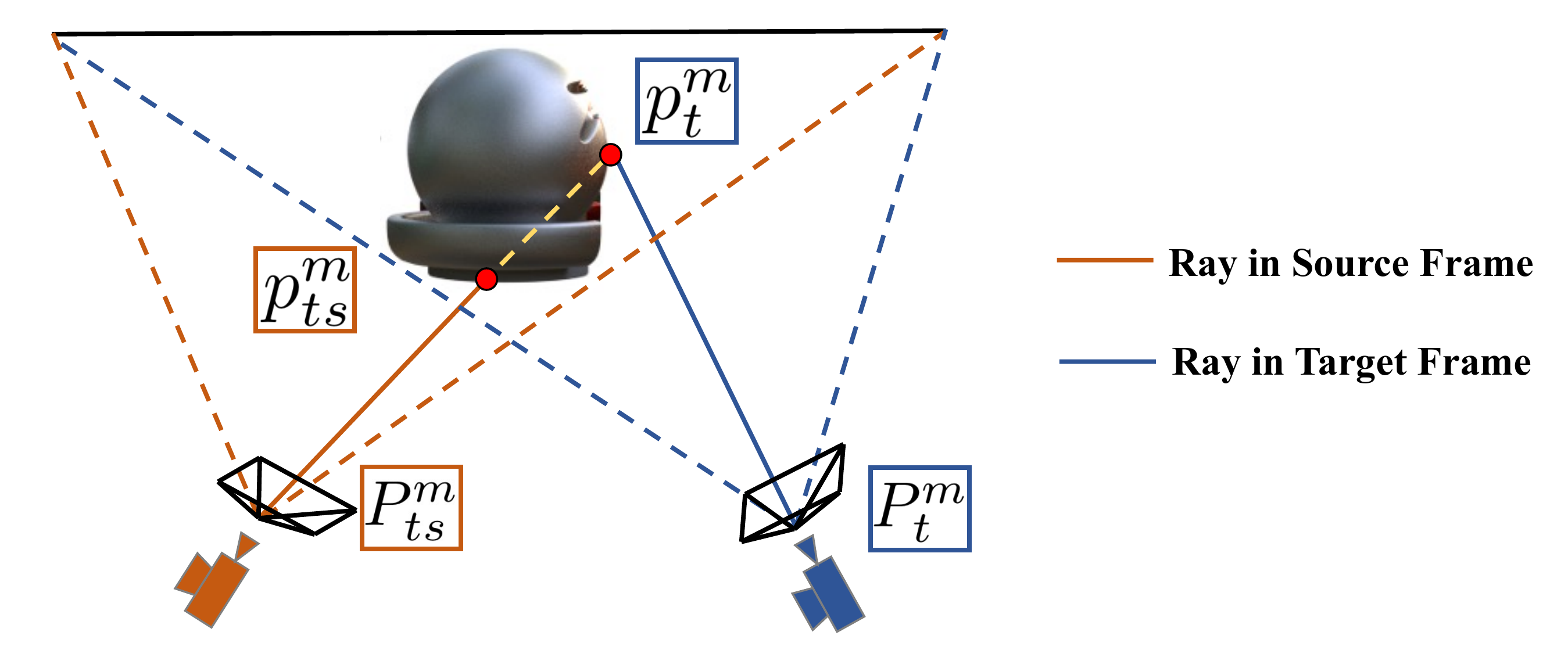}
    \caption{ \textbf{The illustration of the occlusion regularization.}  This figure shows the intuition behind proposed occlusion regularization where camera is moving from left to right.}
\label{fig:fig4}
\end{figure} 

\noindent\textbf{Regularization for Occlusion:} Due to dynamic objects and camera motion, occlusions are inevitable in application scenarios.
These occlusions would cause ambiguities in the photometric warping between source and target frames. To minimize its influence, we design a novel occlusion regularization strategy to mitigate this problem as illustrated in Fig.\ \ref{fig:fig4}. During the training phase, after predicting the depth map $Z_t$ of the target frame, the intersection $p_t^m(x, y, z)$ between the rays of pixel $P_t^m(X, Y, 1)$ and the surfaces in the target camera frustum can be formulated as 
\begin{equation}
    p_t^m = Z_t(P_t^m) K^{-1}_t P_t^m,
\end{equation}
whose depth with respect to the source frame is denoted as $Z_s(p_t^m)$. Then the corresponding pixel $P_{ts}^m$ in the source pixel frame is:
\begin{equation}
    P_{ts}^m = \frac{1}{Z_s(p_t^m)} K_{s}   T_{ts} p_t^m,
\end{equation}
whose depth is denoted as $Z_s(P_{ts}^m)$ and corresponding points $p_{ts}^m$ in the source frame could also be found. Due to the movement of object or the motion of camera, point $p_{ts}^m$ would be different from point $p_t^m$. This difference could be used to compute the occlusion mask, which could be written as:
\begin{equation}
    \mathcal{M}_o = (|Z_s(p_t^m) - Z_s(P_{ts}^m)| < \gamma),
\end{equation}
where $\gamma$ is the threshold in terms of the depth difference. 

\noindent\textbf{Regularization for Brightness:}
With the help of the Eqn.\ \eqref{eqn:warp}, we project the source frame $I_s$ to the target frame and get the reconstructed target image $I_{st}$, which are used in photometric loss function construction introduced in Sec.\ \ref{sec3:self-supervised}. 
However, despite the proposed occlusion regularization, 
the adjacent frames of outdoor scenarios often suffer from illumination inconsistency. Directly forcing reconstructed pixel values to equal to original pixel values will inevitably deteriorate the prediction performance. Hence, we apply the following regularization to transform the reconstructed image $I_{st}$:
\begin{equation}
    I_{st}^{ab} = a_{st} I_{st} + b_{st},
\end{equation}
where $a$ and $b$ are the brightness transformation parameters as in \cite{yang2020d3vo, engel2017direct, engel2015large, jin2001real}. 

\begin{figure*}[t]
\setlength{\abovecaptionskip}{0.1cm}
\setlength{\belowcaptionskip}{-0.8cm}
    \centering
    \includegraphics[width=12.0cm]{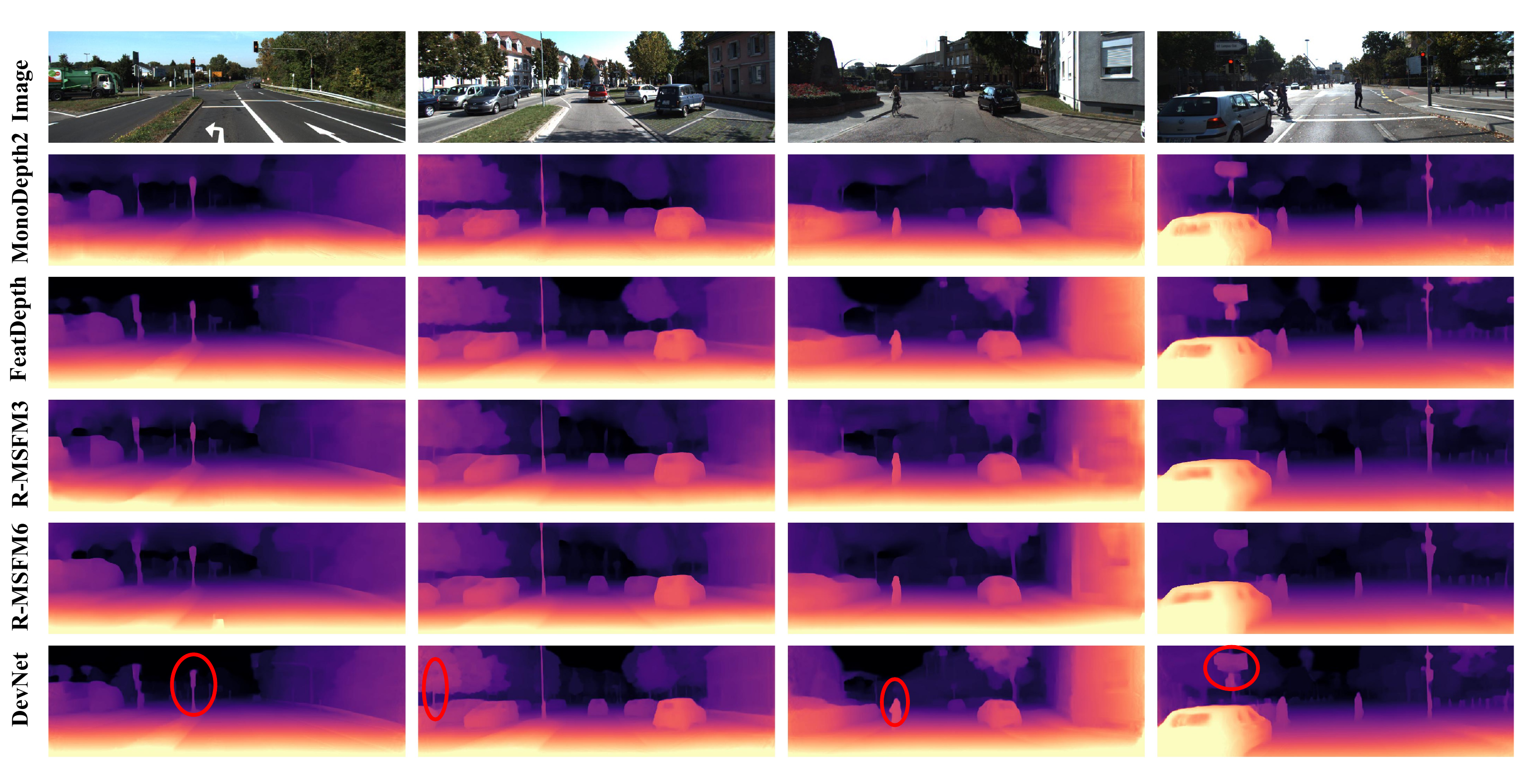}
    \caption{\textbf{The qualitative results of depth estimation on the KITTI-2015.} The results generated by DevNet are shown in the last row, which are sharper than results estimated by listed methods. This comparison is consistent with quantitative results in Tab.\ \ref{table:depth}. }
\label{fig:fig5}
\end{figure*} 

\subsection{Training Loss}\label{sec3:self-supervised}
Our training loss function consists of four parts, including a photometric loss $\mathcal{L}_p$, a smoothness loss $\mathcal{L}_s$, a depth synthesis loss $\mathcal{L}_d$ and a regularization loss $\mathcal{L}_r$:
\begin{equation}
    \mathcal{L} = \mathcal{L}_p + \alpha \mathcal{L}_s + \beta \mathcal{L}_d + \eta \mathcal{L}_r,
\label{eqn:loss_function}
\end{equation}
where $\alpha$, $\beta$ and $\eta$ are corresponding weights to balance these losses. 

\noindent\textbf{The photometric loss function} consists of L1 and SSIM  \cite{godard2017unsupervised, godard2019digging}:
\begin{equation}
\begin{split}
    \mathcal{L}_p & = 0.15 * \sum_{p} \mathcal{M}_o \odot \mathcal{M}_i \odot |I_t - I^{ab}_{st}| \\ &+ 0.85 * \frac{1 - \text{SSIM}(\mathcal{M}_o \odot \mathcal{M}_i \odot I_t, \mathcal{M}_o \odot \mathcal{M}_i \odot I^{ab}_{st})}{2}, 
\end{split}
\label{eqn:loss_photo}
\end{equation}
where SSIM is used to quantify the quality degradation in the image processing \cite{wang2004image} and $\mathcal{M}_i$ is the identity mask provided in MonoDepth2\cite{godard2019digging}.

\noindent\textbf{The smoothness loss function} is applied to the target depth map $Z_t$ as:
\begin{equation}
    \mathcal{L}_s = |\partial_x Z_t^*| e^{- |\partial_x I_t|} + |\partial_y Z_t^*| e^{- |\partial_y I_t|},
\label{eqn:loss_smooth}
\end{equation}
where $Z_t^* = Z_t / \overline{Z_t}$ and $\overline{Z_t}$ is the average inverse depth of target frame \cite{wang2018learning}. This loss function is used to perverse the edge in the depth reconstruction. 

\noindent\textbf{The depth consistency loss.} Based on this Eqn.\ \eqref{eqn:warp}, our DevNet  reconstructs the depth map of target frame $Z_{st}$ based on the density volume of source frame $V^j(\bm{\sigma}_s)$, as the density of each point $\bm{\sigma}_t(p_{t})$ in target frame can be found by querying the density information of its counterpart in source frame $\bm{\sigma}_s(T_{st}p_{t})$.
The depth consistency loss is formulated as:
\begin{equation}
     \mathcal{L}_d = \frac{1}{M'} \sum_{m} ( \frac{|\mathcal{M}_o \odot Z^m_{st} - \mathcal{M}_o \odot Z^m_{t}|}{\mathcal{M}_o \odot Z^m_{st} + \mathcal{M}_o \odot Z^m_{t}}) ,
\label{eqn:depth_loss}
\end{equation}
where $M'$ is the number of estimated valid depth values.

\noindent\textbf{The regularization loss} $\mathcal{L}_r$ regularizes brightness transformation parameters:
\begin{equation}
    \mathcal{L}_r = (a-1)^2 + b^2,
\label{eqn:reg_loss}
\end{equation}
which is proposed under the assumption that there is no significant illumination inconsistency between adjacent frames.

\begin{table*}[hbt!]
\centering
\setlength{\abovecaptionskip}{0.15cm}
\setlength{\belowcaptionskip}{-0.65cm}
\scalebox{0.85}{\begin{tabular}{r|c|cccc|ccc}
\Xhline{2.0\arrayrulewidth}
\multicolumn{1}{c|}{}& \multicolumn{1}{c|}{} & \multicolumn{4}{c|}{Lower is Better}& \multicolumn{3}{c}{Larger is Better}\\ \cline{3-9} 
\multicolumn{1}{r|}{\multirow{-2}{*}{Methods}} & \multicolumn{1}{c|}{\multirow{-2}{*}{Type}} & \multicolumn{1}{l|}{Abs Rel} &
\multicolumn{1}{l|}{Sq Rel} & \multicolumn{1}{l|}{RMSE} & \multicolumn{1}{l|}{RMSE log} & \multicolumn{1}{l|}{$\delta<1.25$} & \multicolumn{1}{l|}{$\delta<1.25^2$} & \multicolumn{1}{l}{$\delta<1.25^3$} \\ 
\Xhline{2.0\arrayrulewidth}
SIGNet\cite{meng2019signet} & M & 0.133 & 0.905 & 5.181  & 0.208 & 0.825 & 0.947 & 0.981 \\ 
LearnK\cite{gordon2019depth} & M & 0.128 & 0.959 & 5.230 & 0.212 & 0.845 & 0.947 & 0.976 \\ 
DualNet\cite{zhou2019unsupervised}& M & 0.121 & 0.837 & 4.945 & 0.197& 0.853 & 0.955 & 0.982 \\ 
MonoDepth2(R50)\cite{godard2019digging} & M& 0.110  & 0.831 & 4.642 & 0.187& 0.883& 0.962 & 0.982 \\ 
Johnston\cite{johnston2020self} & M & 0.106 & 0.861 & 4.699 & 0.185 & 0.889 & 0.962 & 0.982 \\
FeatDepth(R50)\cite{shu2020feature} & M & 0.104 & 0.729 & 4.481 & 0.179 & \textbf{0.893} & 0.965 &  0.984 \\ 
PackNet-SfM\cite{guizilini20203d} & M & 0.107 & 0.802 & 4.538 & 0.186 & 0.889 & 0.962 & 0.981\\ 
R-MSFM6\cite{zhou2021r} & M & 0.108 & 0.748 & 4.470 & 0.185 & 0.889 & 0.963 & 0.982\\ 
\textbf{w/o Pretrain(R50)(P24)} & M & 0.114 & 0.836 & 4.888 & 0.191 & 0.877 & 0.958 & 0.981 \\ 
\textbf{DevNet(R50)(P16)} & M & 0.103 & 0.713 & 4.459 & 0.177 & 0.890 & 0.965 & 0.982 \\ 
\textbf{DevNet(R50)(P24)} & M & \textbf{0.100} & \textbf{0.699} & \textbf{4.412} & \textbf{0.174} & \textbf{0.893} & \textbf{0.966} & \textbf{0.985}  \\ 
\hdashline
DFR\cite{zhan2018unsupervised} & MS & 0.135 & 1.132 & 5.585 & 0.229 & 0.820 & 0.933 & 0.971 \\ 
EPC++\cite{luo2019every}& MS & 0.128 & 0.935 & 5.011 & 0.209 & 0.831 & 0.945 & 0.979\\ 
MonoDepth2(R50)\cite{godard2019digging}& MS & 0.106 & 0.818 & 4.75 & 0.196 & 0.874 & 0.957& 0.979  \\ 
FeatDepth(R50)\cite{shu2020feature} & MS & 0.099 & 0.697 & 4.427 & 0.184 & 0.889 & 0.963 & 0.982 \\ 
R-MSFM6\cite{zhou2021r} & MS & 0.108 & 0.753 & 4.469 & 0.185 & 0.888 & 0.963 & 0.982 \\ 
\textbf{w/o Pretrain(R50)(P24)} & MS & 0.107 & 0.751 & 4.461 & 0.189 & 0.883 & 0.963 & 0.981 \\ 
\textbf{DevNet(R50)(P24)} & MS & \textbf{0.095} & \textbf{0.671} & \textbf{4.365} & \textbf{0.174} & \textbf{0.895} & \textbf{0.970}  & \textbf{0.988} \\ 
\hdashline
NeWCRFs\cite{yuan2022new} & MSup & 0.052 & 0.155 & 2.129 & 0.079 & 0.974 & 0.997 & 0.999 \\ \Xhline{2.0\arrayrulewidth}
\end{tabular}}
\caption{\textbf{The quantitative results of depth estimation on the KITTI-2015.} Comparison between DevNet and other depth estimation methods on KITTI-2015 with the Eigen split. Best results are in bold. M: trained with monocular sequence.  MS: trained with both monocular sequence and stereo images. MSup: supervised trained with monocular sequence.}
\label{table:depth}
\end{table*}

\subsection{Discussions}
\label{sec3:discussion}
\textbf{The Relation with NeRF\cite{mildenhall2020nerf}}
\textbf{Similarity:} 
Both of us reconstruct the color image or the depth map by discretizing the scenario into thousands of points. The color or depth value of each pixel is calculated by integrating along with its rays. \textbf{Difference:} (1) Instead of running the neural network \textit{thousands of times} to obtain point-wise information for the whole scene, DevNet only passes through the network \textit{once} to acquire the density information of the whole scene. (2) DevNet concentrates on self-supervised depth estimation and this network can be used in multiple scenarios, while NeRF focuses on learning an implicit representation 
in a certain scenario. 

\noindent\textbf{The Relation with FLA \cite{gonzalezbello2020forget}}
\textbf{Similarity:} 
Both of us adapt the auto-encoder structure and output multi-channel geometry information used for depth map construction. \textbf{Difference:} (1) FLA outputs multi-channel disparity logit volume which constructs depth through \textit{SoftMax} calculation. DevNet outputs multi-channel density volume which generates depth through \textit{integration}. (2) FLA pre-defines the position of each channel and does not consider the position information. At each time, DevNet randomly takes samples from a uniform distribution, and the position is used to construct embedded depth volume $V_i^j$. (3)
Unlike FLA, DevNet does not need the two-stage training strategy and realizes end-to-end training. (4) DevNet focuses on self-supervised \textit{monocular} depth learning, while FLA focuses on self-supervised $stereo$ depth learning.

\section{Experiments}\label{sec:exp}
Our experiments are mainly conducted on the KITTI-2015 \cite{geiger2012we} and NYU-V2 \cite{silberman2012indoor}. For KITTI-2015, the Eigen split  \cite{eigen2015predicting} is used to train and test our model. For a fair comparison, following the same data selection strategy in previous research \cite{zhou2017unsupervised, zhou2021r}, we remove all static images from the training set, after which 39,810 training images and 4,424 test images are generated. We adapt the camera matrix setting in \cite{godard2019digging}, which uses the same intrinsic matrix for all images. The length focal is set as the average value. For NYU-V2, we adapt the official train and test split used in \cite{zhao2020towards}, where the train set contains 20000 images and the test set consists of 654 labeled images. In this section, the performance of DevNet is firstly tested in terms of depth prediction. Depth prediction results are compared with following widely used evaluation metrics \cite{godard2019digging, eigen2014depth}:

\begin{itemize}
    \item $\text{Abs}\ \text{Rel} = \frac{1}{|M|} \sum^M_{m=1} \frac{|\hat{Z}^m - Z^m|}{\hat{Z}^m}$; 
    \item $\text{Sq}\ \text{Rel} = \frac{1}{|M|} \sum^M_{m=1} \frac{||\hat{Z}^m - Z^m||_2}{\hat{Z}^m}$; 
    \item $\text{RMSE} = \sqrt{\frac{1}{|M|} \sum^M_{m=1} ||\hat{Z}^m - Z^m||_2}$;
    \item $\text{RMSE}\ \log= \sqrt{\frac{1}{|M|} \sum^M_{m=1} ||log (\hat{Z}^m) - log(Z^m)||_2}$;
    \item $\frac{1}{M}|\{\delta = \max(\frac{\hat{Z}^m}{Z^m}, \frac{Z^m}{\hat{Z}^m}, ) < 1.25^t, m=1,...,M\}|$;
\end{itemize}
where $\hat{Z}^m$ is the ground truth; $Z^m$ is predicted depth value and $t = 1, 2, 3$. Then, to evaluate each component in DevNet, ablation studies are also conducted by varying the sample number, removing mask and loss functions, and changing the backbone. Besides the depth estimation, the DevNet is capable of achieving visual odometry estimation, which is compared to baselines. Finally, the generalization of DevNet trained on KITTI-2015 is directly tested on the NYU-V2 \cite{silberman2012indoor}. 

\begin{table}[hbt!]
\centering
\setlength{\abovecaptionskip}{0.15cm}
\setlength{\belowcaptionskip}{-0.95cm}
\scalebox{0.85}{\begin{tabular}{c|c|ccccccc}
\Xhline{2.0\arrayrulewidth}
\multicolumn{1}{c|}{Model} &
\multicolumn{1}{c|}{Type} &\multicolumn{1}{c|}{Abs Rel} & \multicolumn{1}{c|}{Sq Rel} & \multicolumn{1}{c|}{RMSE} & \multicolumn{1}{c|}{RMSE log} & \multicolumn{1}{c|}{$\delta < 1.25$}  & \multicolumn{1}{c|}{$\delta < 1.25^2$}  & \multicolumn{1}{c}{$\delta < 1.25^3$}\\ 
\Xhline{2.0\arrayrulewidth}
MonoDepth2(R18)\cite{godard2019digging} & M & 0.160 & 0.152 & 0.601 & 0.186 & 0.767 & 0.949 & 0.988\\ 
Monoindoor(R18)\cite{ji2021monoindoor} & M & 0.134 & - & 0.526 & - & 0.823 & 0.958 & 0.989\\
\textbf{DevNet(R18)(P24)} & M & \textbf{0.129} & \textbf{0.098} & \textbf{0.492} & \textbf{0.153} & \textbf{0.843} & \textbf{0.965} & \textbf{0.992} \\     
\Xhline{2.0\arrayrulewidth}
\end{tabular}}
\caption{\textbf{The quantitative results of depth estimation on the NYU-V2.} Comparison between DevNet and other depth estimation methods on NYU-V2 with the official split. Best results are in bold. }
\label{table:NYU1}
\end{table}

\subsection{Depth Estimation}
\noindent\textbf{KITTI-2015:} In this experiment, the depth encoder adapts ResNet50(R50), while the pose encoder adapts ResNet18(R18) as base model. During the training phase, our DevNet takes the image with the resolution of $320\times1024$ as input. When testing our DevNet, we set $80m$ as the max depth value and $0.1m$ as the minimum depth value \cite{godard2019digging, jin2001real}. Same as in \cite{zhou2017unsupervised}, the median value of the ground-truth image is used to calculate the scale factor. The comparison between the performance of DevNet and that of state-of-art algorithms is demonstrated in Tab.\ \ref{table:depth}. It is clear to see that DevNet generally performs better no matter using monocular images, or a combination of monocular and stereo images. Compared to \cite{johnston2020self} that also learns a disparity volume used to generate depth map through SoftMax, DevNet outperforms them largely by $5.7\%$ in terms of $\text{Abs \ Rel}$. FeatDepth \cite{shu2020feature} proposes to pre-train an feature encoder for low-texture region learning. DevNet outperforms them by $3.8\%$ in terms of $\text{Abs\ Rel}$ without pretraining the feature auto-encoder. A qualitative comparison of depth predictions is presented in Fig.\ \ref{fig:fig5}, showing that DevNet detects the details of moving objects and thin objects, highlighted by red circles. One thing worth mentioning, even though DevNet could achieve good performance with unsupervised setting, supervised algorithms, e.g., NeWCRFs\cite{yuan2022new}, still largely outperform unsupervised algorithms. $Size \& Speed:$ One thing worth mentioning is that DevNet slightly increases the parameter size and the running time only, as DevNet only modifies skip-connects and the last layer channel number of MonoDepth2 (R50) backbone. While DevNet significantly improves the performance.  
\newline
\noindent\textbf{NYU-V2:} For this indoor dataset, during the training phase, DevNet is fed with a resolution of $288\times384$. We set $10.0m$ as the max depth value and $0.1m$ as the minimum depth value. Considering the lower resolution of NYU-V2, we adapt the ResNet18 backbone for both depth and pose modules. The result in Tab.\ \ref{table:NYU1} shows that DevNet also outperforms its counterparts in the indoor dataset. Specifically, DevNet outperforms Monoindoor by around $4\%$ in terms of $\text{Abs\ Rel}$. 

\begin{table}[]
\centering
\setlength{\abovecaptionskip}{0.15cm}
\setlength{\belowcaptionskip}{-0.9cm}
\scalebox{0.85}{\begin{tabular}{r|cc|cc}
\Xhline{2.0\arrayrulewidth}
\multicolumn{1}{c|}{} & \multicolumn{2}{c|}{Seq9} & \multicolumn{2}{c}{Seq10} \\ \cline{2-5} 
\multicolumn{1}{c|}{\multirow{-2}{*}{Algorithm}} & \multicolumn{1}{l|}{Translation} & \multicolumn{1}{l|}{Rotation} & \multicolumn{1}{l|}{Translation} & \multicolumn{1}{l}{Rotation} \\ 
\Xhline{2.0\arrayrulewidth}
DFR \cite{zhan2018unsupervised} & 11.93 & 3.91 & 12.45 & 3.46 \\ 
MonoDepth2 \cite{godard2019digging} & 10.85 & 2.86 & 11.60 & 5.72 \\ 
NeuralBundler \cite{li2019pose} & 8.10  & 2.81 & 12.90 & \textbf{3.71} \\ 
SC-SfMlearner \cite{zhou2017unsupervised} & 8.24 & 2.19 & 10.70 & 4.58 \\ 
FeatDepth \cite{shu2020feature} & 8.75 & 2.11 & 10.67 & 4.91 \\ 
\textbf{DevNet(R50)(P24)} & \textbf{8.09} & \textbf{2.05}  & \textbf{10.05} & 4.32 \\ \Xhline{2.0\arrayrulewidth}
\end{tabular}}
\caption{\textbf{The quantitative results of odometry estimation on the KITTI odometry dataset.} Comparison between DevNet and others in terms of Average translational root mean square error drift and average rotational root mean square error drift.}
\label{tab:odo}
\end{table}

\subsection{Odometry Estimation}
To test the performance of odometry estimation, we adapt the split strategy used in \cite{zhou2017unsupervised, zhan2018unsupervised}. This strategy uses the Sequences 00-08 of the KITTI odometry dataset as the training set and the Sequences 09-10 as the test set, as shown in Tab.\ \ref{tab:odo}. The averaged translational and rotational root mean square error (RMSE) are reported. Our DevNet outperforms other representative learning-based methods, e.g. MonoDepth2 and FeatDepth. This is because our DevNet focuses on learning the geometric correspondences between adjacent frames and its accurate depth estimation benefits odometry estimation when jointly optimizing both the odometry module and depth estimation module.

\begin{table}[]
\centering
\setlength{\abovecaptionskip}{0.15cm}
\setlength{\belowcaptionskip}{-0.65cm}
\scalebox{0.85}{\begin{tabular}{c|c|cccccc}
\Xhline{2.0\arrayrulewidth}
\multicolumn{1}{c|}{Regularization} & \multicolumn{1}{c|}{Type}  & \multicolumn{1}{c|}{Abs Rel} & \multicolumn{1}{c|}{Sq Rel} & \multicolumn{1}{c|}{RMSE} & \multicolumn{1}{c|}{$\delta < 1.25$} & \multicolumn{1}{c|}{$\delta < 1.25^2$} & \multicolumn{1}{c}{$\delta < 1.25^3$} \\ 
\Xhline{2.0\arrayrulewidth}
Baseline(R50)(P24) & M & 0.118 & 0.881 & 4.897 & 0.872 & 0.950 & 0.971\\ 
w/o Embedding(R50)(P24) & M & 0.102 & 0.711 & 4.439 & 0.889 & 0.963 & 0.983 \\ 
w/o Color Reg(R50)(P24) & M & 0.103 & 0.719 & 4.462 & 0.889 & 0.962 & 0.979\\ 
w/o Occlusion Mask(R50)(P24) & M & 0.106 & 0.791 & 4.547 & 0.882 & 0.959 & 0.978 \\ 
w/o Depth Loss(R50)(P24) & M & 0.107 & 0.784 & 4.487 & 0.880 & 0.959 & 0.967 \\ 
DevNet(R18)(P24) & M & 0.108 & 0.801 & 4.712 & 0.874 & 0.951 & 0.973 \\
DevNet(R50)(P08) & M & 0.108 & 0.746 & 4.471 & 0.884 & 0.949 & 0.969 \\
DevNet(R50)(P16) & M & 0.103 & 0.713 & 4.459 & 0.890 & 0.965 & 0.982 \\ 
DevNet(R50)(P24) & M & 0.100 & 0.699 & 4.412 & \textbf{0.893} & 0.966 & 0.985 \\ 
DevNet(R50)(P32) & M & \textbf{0.100} & \textbf{0.697} & \textbf{4.408} & 0.891  & \textbf{0.968} & \textbf{0.988} \\ 
\Xhline{2.0\arrayrulewidth}
\end{tabular}}
\caption{\textbf{The ablation study on depth estimation above the KITTI-2015.} The ablation studies concern the influences of different components, training strategies, and different sample number of planes.}
\label{table:ablation1}
\end{table}

\subsection{Ablation Study}
In this section, we analyze the performance of DevNet by studying the influence of different components and training strategies on the depth prediction result.

\noindent\textbf{Training Strategies:} To evaluate the contribution of each training strtegy towards our DevNet framework, we conducted ablation studies under several settings, including the DevNet without embedding strategy applied on position information, without color regularization, without occlusion mask, and without depth loss. 
The baseline algorithm denotes the DevNet with only photometric loss, smooth loss, and general min-reprojection mask. Based on the analysis in Sec.\ \ref{sec:method}, adding the proposed components in the training process is useful to improving estimation accuracy, which is also verified in Tab.\ \ref{table:ablation1}. Hence, the main concern here is which component contributes largest towards the model. By comparing the results of DevNet without certain components, it indicates that the position embedding strategy and the color regularization are less influential compared with the occlusion mask and depth loss. The performances of DevNet w/o occlusion mask and DevNet w/o depth mask are similar, as both of two terms focus on the spatial correspondence. However, the highest performance achieved by DevNet shows that the occlusion mask and the depth loss are beneficial towards each other. \noindent\textbf{Small Backbone:} Resnet50 is used as our backbone. To study the influence of backbone with different size, we also test the performance DevNet with Resnet18 in Tab.~\ref{table:ablation1}. We could notice that replacing the backbone R50 with the R18 could decline
DevNet’s performance, while DevNet still outperforms the presented methods on KITTI-2015. \textbf{Sampled Planes Number:} DevNet splits the camera frustum into multiple parallel planes, therefore we are also interested in the influence of the number of plane samples towards the performance of DevNet. As demonstrated in Tab.\ \ref{table:ablation1}, increasing the number of samples improves the estimation accuracy, which is noticed by comparing the performance of DevNet(P08), DevNet(P16), DevNet(P24) and DevNet(P32). However, it can be observed that when the number of samples is already large enough, increasing the number of samples will not result in a high margin in terms of prediction accuracy.  

\subsection{Generalization}
To test the generalization ability of our DevNet, we adapt the DevNet model trained on the KITTI-2015, to directly evaluate it above the NYU-V2 \cite{silberman2012indoor} without fine-tuning shown in Tab.\ \ref{tab.nyu} and Fig.\ \ref{fig:fig7}. Despite huge differences, indoor and outdoor scenes still share some features, e.g., color inconsistency probably indicates depth difference, which makes this generalization possible. The network is fed with images with a resolution of $288\times384$. The comparison among FeatDepth, R-MSFM3, and R-MSFM6 indicates that better performance on the KITTI-2015 doesn't necessarily mean better performance on the NYU-V2. It also could be noticed that our model outperforms other monocular self-supervised methods in all evaluation metrics, which is a shred of strong evidence to support that our model has better generalization capability and could effectively learn the geometric information based on appearance information. 

\begin{table}[]
\centering
\setlength{\abovecaptionskip}{0.15cm}
\setlength{\belowcaptionskip}{-0.99cm}
\scalebox{0.85}{\begin{tabular}{r|c|cccccc}
\Xhline{2.0\arrayrulewidth}
Algorithm  & Type & \multicolumn{1}{l|}{Abs    Rel} & \multicolumn{1}{c|}{Sq Rel} & \multicolumn{1}{c|}{RMSE} & \multicolumn{1}{c|}{$\delta < 1.25$} & \multicolumn{1}{c|}{$\delta < 1.25^2$} & \multicolumn{1}{c}{$\delta < 1.25^3$} \\ 
\Xhline{2.0\arrayrulewidth}
MonoDepth2 \cite{godard2019digging} & M & 0.391 & 0.898 & 1.458 & 0.415 & 0.724 & 0.874 \\ 
FeatDepth \cite{shu2020feature} & M & 0.353 & 0.701 & 1.231 & 0.516 & 0.776 & 0.904\\ 
R-MSFM3 \cite{zhou2021r} & M & 0.355 & 0.672 & 1.276 & 0.476 & 0.755 & 0.898\\ 
R-MSFM6 \cite{zhou2021r} & M & 0.372 & 0.803 & 1.344 & 0.494 & 0.746 & 0.884\\ 
DevNet(R50)(P24)  & M & \textbf{0.333} & \textbf{0.605} & \textbf{1.142} & \textbf{0.541} & \textbf{0.790} & \textbf{0.911}\\
\Xhline{2.0\arrayrulewidth}
\end{tabular}}
\caption{\textbf{The quantitative results of depth estimation on NYU-V2.} Algorithms are trained on monocular video of KITTI-2015 and tested on NYU-V2. }
\label{tab.nyu}
\end{table}

\begin{figure}[t]
\setlength{\belowcaptionskip}{-0.65cm}
    \centering
    \includegraphics[width=10.5cm]{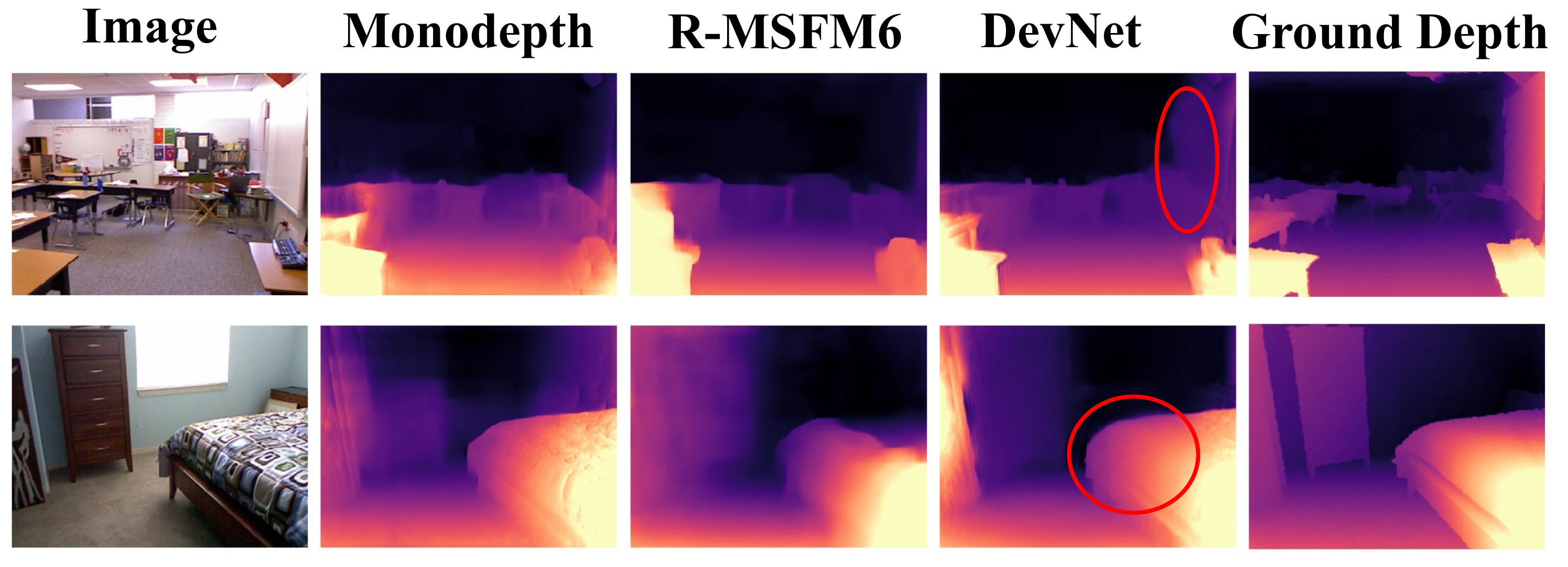}
    \caption{\textbf{The qualitative results of depth estimation on the NYU-V2.} All methods are trained on KITTI-2015 and tested on NYU-V2.}
\label{fig:fig7}
\end{figure} 

\section{Conclusion}
In this work, we propose DevNet, an effective self-supervised monocular depth learning framework, that can accurately predict depth maps by constructing the density volume of the scene. 
To mitigate the influences of occlusions and illumination inconsistency, a novel occlusion mask and a brightness regularization are designed to benefit the training process of this framework. Moreover, the depth consistency loss function is designed to maintain the consistency of density volume for successive frames in the temporal domain. We show that using the training strategy consisting of them provides a simple and efficient model to produce monocular depth and odometry estimation accurately for both indoor and outdoor scenarios, which also possesses high generalization ability.

\smallskip\noindent\textbf{Acknowledgements.} This work was partially supported by the Huawei UK AI Fellowship. We gratefully acknowledge the support of MindSpore, CANN(Compute Architecture for 
Neural Networks) and Ascend AI Processor used for this research.
\clearpage
%
%
\bibliographystyle{plain}
\bibliography{egbib}
\end{document}